\documentclass[runningheads]{llncs}
\usepackage{graphicx}

\usepackage{tikz}
\usepackage{comment}
\usepackage{amsmath,amssymb} 
\usepackage{color}
\usepackage{multirow}
\usepackage{orcidlink}

\usepackage[utf8]{inputenc}
\usepackage{amsmath}
\usepackage{amssymb}
\usepackage{hyperref}
\usepackage{outlines}
\usepackage{array,graphicx}
\usepackage{booktabs}
\usepackage{pifont}
\usepackage{tabularx}
\usepackage{subcaption}
\usepackage[title]{appendix}
\usepackage{adjustbox}

\definecolor{OliveGreen}{rgb}{0,0.6,0}
\definecolor{Red}{rgb}{1,0,0}

\definecolor{textblue}{rgb}{.2,.2,.7}
\definecolor{textred}{rgb}{0.54,0,0}
\definecolor{textgreen}{rgb}{0,0.43,0}
\usepackage{listings}

\newcommand*\rot{\rotatebox{90}}
\newcommand*\OK{\textcolor{OliveGreen}{\ding{51}}}
\newcommand*\NO{\textcolor{Red}{\ding{55}}}

\usepackage{xifthen}
\usepackage{xparse}
\usepackage{subdepth}
\usepackage{leftidx}
\usepackage{bm}

\newcommand{\bbm}{\begin{bmatrix}}
\newcommand{\ebm}{\end{bmatrix}}

\DeclareMathAlphabet{\mbf}{OT1}{ptm}{b}{n}
\newcommand{\mbs}[1]{{\bm{#1}}}

\newcommand{\mbsbar}[1]{{\overline{\boldsymbol{#1}}}}
\newcommand{\mbshat}[1]{{\hat{\boldsymbol{#1}}}}
\newcommand{\mbstilde}[1]{{\tilde{\boldsymbol{#1}}}}

\newcommand{\mbsdot}[1]{{\dot {\boldsymbol{#1}}}}

\newcommand{\mbfbar}[1]{{\overline{\mbf{#1}}}}
\newcommand{\mbfhat}[1]{{\hat{\mbf{#1}}}}
\newcommand{\mbftilde}[1]{{\tilde{\mbf{#1}}}}

\newcommand{\mbfdot}[1]{{\dot{\mbf{#1}}}}

\newcommand{\cframe}[1]{{\smash{\protect\underrightarrow{\mathcal{F}}_{#1}}}}

\DeclareMathAlphabet{\mathbfit}{OML}{cmm}{b}{it}
\newcommand{\homo}[1]{{\mathbfit{#1}}}

\newcommand{\mbfhtilde}[1]{{\tilde{\homo{#1}}}}

\newcommand{\mbfh}[1]{{\homo{#1}}}

\newcommand{\pos}[2]{\leftidx{_{#1}}{ \mbf r}{_{#2}}} 
\newcommand{\postilde}[2]{\leftidx{_{#1}}{\mbftilde r}{_{#2}}} 

\newcommand{\rotvec}[2]{\leftidx{_{#1}}{ \mbs \alpha }{_{#2}}}

\newcommand{\lmh}[1]{\leftidx{_{#1}}{\mbfh l}{}} 

\newcommand{\vel}[3]{\leftidx{_{#1}}{\mbf v}{\IfValueTF{#2}{_{#2#3\hspace{2pt}}}{}}} 
\newcommand{\veltilde}[3]{\leftidx{_{#1}}{\mbftilde v}{\IfValueTF{#2}{_{#2#3\hspace{2pt}}}{}}} 
\newcommand{\velbar}[3]{\leftidx{_{#1}}{\mbfbar v}{\IfValueTF{#2}{_{#2#3\hspace{2pt}}}{}}} 
\newcommand{\velhat}[3]{\leftidx{_{#1}}{\mbfhat v}{\IfValueTF{#2}{_{#2#3\hspace{2pt}}}{}}} 
\newcommand{\veldot}[3]{\leftidx{_{#1}}{\mbfdot v}{\IfValueTF{#2}{_{#2#3\hspace{2pt}}}{}}} 

\newcommand{\myvec}[2]{\leftidx{_{#2}\hspace{-1pt}}{\mbf #1}{}} 

\newcommand{\acc}[3]{\leftidx{_{#1}}{\mbf a}{\IfValueTF{#2}{_{#2#3\hspace{2pt}}}{}}} 
\newcommand{\acctilde}[3]{\leftidx{_{#1}}{\mbftilde a}{\IfValueTF{#2}{_{#2#3\hspace{2pt}}}{}}} 
\newcommand{\accbar}[3]{\leftidx{_{#1}}{\mbfbar a}{\IfValueTF{#2}{_{#2#3\hspace{2pt}}}{}}} 

\newcommand{\rotvel}[3]{\leftidx{_{#1}}{\mbs \omega}{\IfValueTF{#2}{_{#2#3\hspace{2pt}}}{}}} 
\newcommand{\rotveltilde}[3]{\leftidx{_{#1}}{\mbstilde \omega}{\IfValueTF{#2}{_{#2#3\hspace{2pt}}}{}}} 
\newcommand{\rotvelbar}[3]{\leftidx{_{#1}}{\mbsbar \omega}{\IfValueTF{#2}{_{#2#3\hspace{2pt}}}{}}} 
\newcommand{\rotvelhat}[3]{\leftidx{_{#1}}{\mbshat \omega}{\IfValueTF{#2}{_{#2#3\hspace{2pt}}}{}}} 
\newcommand{\rotveldot}[3]{\leftidx{_{#1}}{\mbsdot \omega}{\IfValueTF{#2}{_{#2#3\hspace{2pt}}}{}}} 

\newcommand{\C}[2]{\leftidx{}{\mbf C}{_{#1#2\hspace{2pt}}}} 
\newcommand{\T}[2]{\leftidx{}{\mbfh T}{_{#1#2\hspace{2pt}}}} 

\newcommand{\real}{\mathbb{R}}
\newcommand{\et}{\emph{et al.} }

\usepackage[font=small,labelfont=bf,width=\textwidth]{caption}
\begin{document}
\pagestyle{headings}
\mainmatter
\def\ECCVSubNumber{7926}

\title{BodySLAM: Joint Camera Localisation, Mapping, and Human Motion Tracking}

\titlerunning{BodySLAM}
\author{
Dorian F. Henning\inst{1}\orcidlink{0000-0003-0449-7053}\index{Henning, Dorian F.}
\and Tristan Laidlow\inst{1}\orcidlink{0000-0002-3487-430X} 
\and Stefan Leutenegger\inst{1,2}\orcidlink{0000-0002-7998-3737}
\thanks{Research presented in this paper has been supported by Dyson Technology Ltd. and the Technical University of Munich}
}
\authorrunning{D. F. Henning et al.}
%
\institute{Imperial College London, UK\\
\email{\{d.henning,t.laidlow15\}@imperial.ac.uk}\\
\and
Technische Universit\"at M\"unchen, Germany\\
\email{stefan.leutenegger@tum.de}}

\maketitle

\begin{abstract}
Estimating human motion from video is an active research area due to its many potential applications.
Most state-of-the-art methods predict human shape and posture estimates for individual images and do not leverage the temporal information available in video.
Many ``in the wild" sequences of human motion are captured by a moving camera, which adds the complication of conflated camera and human motion to the estimation.
We therefore present BodySLAM, a monocular SLAM system that jointly estimates the position, shape, and posture of human bodies, as well as the camera trajectory.
We also introduce a novel human motion model to constrain sequential body postures and observe the scale of the scene.
Through a series of experiments on video sequences of human motion captured by a moving monocular camera, we demonstrate that BodySLAM improves estimates of all human body parameters and camera poses when compared to estimating these separately.

\end{abstract}
\begin{figure}[!htb]
  \begin{minipage}[c]{0.5\textwidth}
    \includegraphics[width=\textwidth]{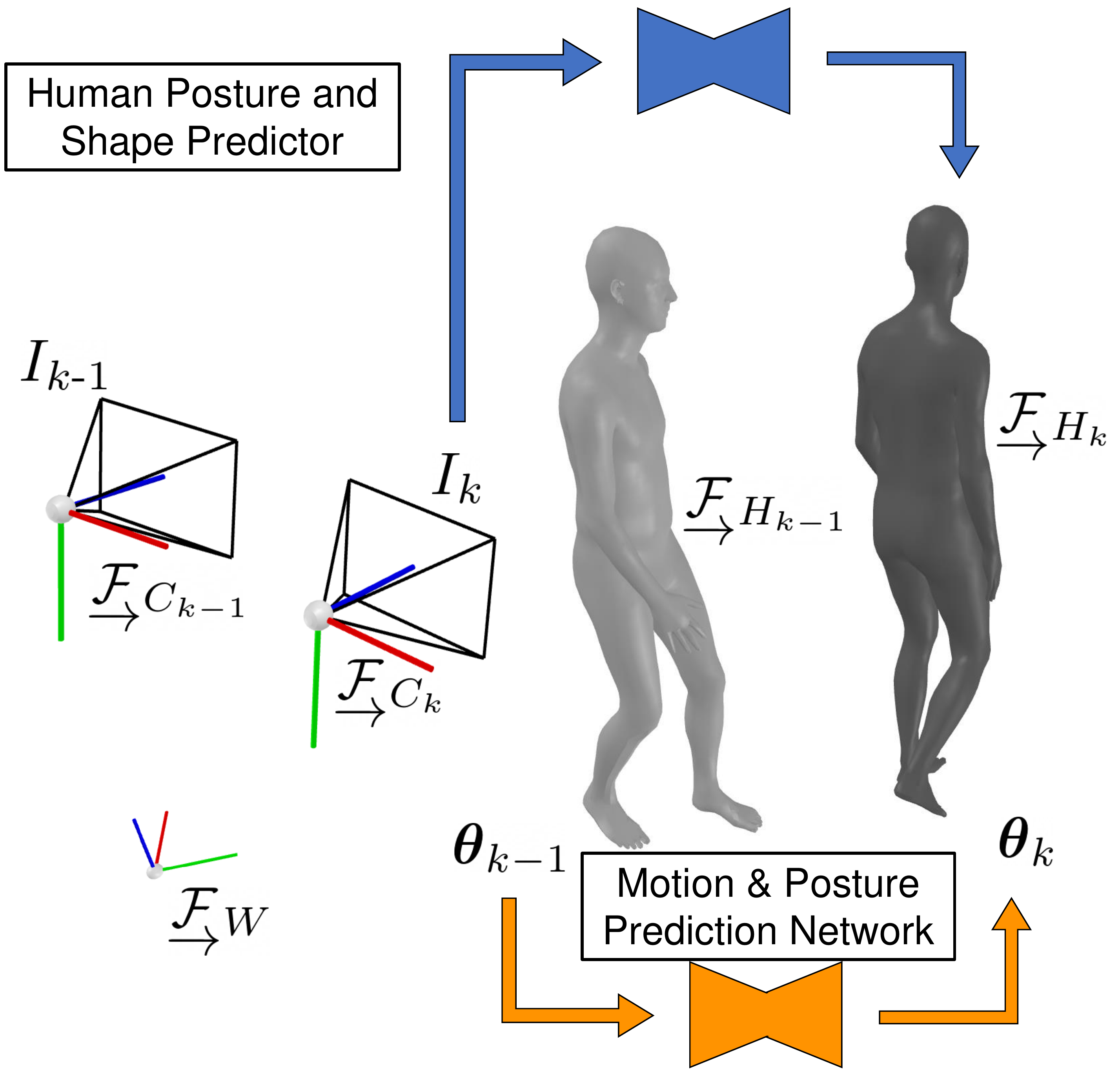}
  \end{minipage}\hfill
  \begin{minipage}[c]{0.5\textwidth}
    \captionsetup{width=0.9\textwidth}
    \caption{BodySLAM Problem Definition and Example Scene. The human body postures of the previous $n$ frames are used to predict its motion; and a Human Mesh Regressor \cite{kolotouros2019spin} is used for human centre pose, posture, and shape prediction. Then, a factor graph of camera trajectory, human states and natural landmarks are jointly optimised.}
    \label{fig:problem_3d}
  \end{minipage}
\end{figure}

\section{Introduction}
Estimating the orientation and translation, shape, and posture of human bodies from a sequence of images is an active research area in computer vision due to its many potential applications in domains such as safe human-robot interaction, biomechanical analysis, action recognition, and augmented and virtual reality (AR/VR).
To fully realise these types of applications at scale, it is desirable that this estimation can be done ``in the wild" from images captured by a moving monocular camera without relying on any special markers or motion-capture systems.
This would even enable human motion analysis on image sequences that were not captured with these applications in mind, such as from videos found on YouTube or other online repositories.
State-of-the-art estimation methods for human shape and posture involve training powerful deep neural networks (DNNs), but these methods require large amounts of annotated training data.
As 3D manual annotations are extremely difficult to obtain, in-the-wild datasets are only labelled in 2D, with 3D annotations only being available in indoor settings where a motion capture system was used to produce ground truth labels.
These restrictions have led to most methods only providing shape and posture estimates for individual frames and not leveraging the temporal information that is available from image sequences.
Another critical limitation is that most of these training datasets only feature stationary cameras, and a true in-the-wild system will need to operate in environments where the camera is moving along an unknown trajectory.
This further complicates the estimation procedure, as with a moving camera, the camera motion and human motion become conflated.

The standard method to incrementally estimate the position of a moving monocular camera in real time is with a visual simultaneous localisation and mapping (SLAM) system.
While many visual SLAM methods have been shown to be both accurate and robust, image sequences containing people in motion pose a particular challenge, since most state-of-the-art visual SLAM methods assume a static environment.
It is possible to use a separate network to segment and mask-out humans from the visual SLAM system, but in many sequences humans take up a large proportion of the frame leaving too few points for the SLAM system to track robustly.
Importantly, even in static scenes with many observable points, classic monocular SLAM systems are not able to observe the scale of a scene, introducing ambiguity to the absolute poses of the cameras and any human motion observed by them.

With these limitations in mind, we propose a method, called BodySLAM, for jointly estimating the orientation and translation, shape, and posture of human bodies, as well as the trajectory of the moving monocular camera that observes them, in a single factor graph optimisation.
We represent 3D human bodies using the shape and posture parameters of the SMPL mesh model \cite{Loper2015SMPL:Model}, and the orientation and translation of the root of the corresponding kinematic tree underlying the SMPL model.
In each frame, we obtain a measurement of these human body parameters using an off-the-shelf neural network.
The body shape measurements are fused together in the factor graph optimisation as these should remain constant for a single individual over the image sequence.
This constraint also helps to resolve the scale ambiguity from the monocular SLAM formulation.

Posture parameters are constrained between frames through the use of a novel motion model.
The motion model uses body shape parameters and estimates of the human centre poses and postures from previous frames to form a prior on future body parameters.
Unlike other models, we choose not to use a recurrent neural network architecture to allow for a simpler factor graph formulation and to enable the marginalisation of previous motions in future work.

Finally, as is standard in visual SLAM, we track and map a sparse set of feature-based landmarks in the background of the image sequences to help estimate the relative 6D camera transformations.
The BodySLAM problem definition for an example scene is provided in Fig.~\ref{fig:problem_3d}.

We demonstrate through a series of quantitative experiments on image sequences featuring human motion and a moving camera that our approach of joint optimisation results in improvements to the estimates of all human body parameters and camera poses when compared to estimating these separately.\footnote{We encourage the reader to view our supplementary video available at:\\ \url{https://youtu.be/0-SL3VeWEvU}}

In summary, the key contributions of this paper are:
\begin{itemize}
    \item a monocular SLAM system, called BodySLAM, that jointly optimises human shape parameters, posture parameters, body centre poses, camera poses, and a set of 3D landmarks,
    \item a novel motion model that robustly predicts future human body parameters from previous ones, and
    \item an experimental evaluation of our joint estimation approach on our own dataset consisting of more than 8k frames with ground truth human body parameters and camera poses obtained by a Vicon motion tracking system, demonstrating that we can accurately recover human body parameters and camera trajectories at metric scale.
\end{itemize}

\section{Related Work}
\subsection{3D Human Body Representations}

3D representations for the human body generally fall into three broad categories: sparse sets of independent body joints, articulated body skeletons with rigidity constraints between joints, and (parametric) human mesh models.

Both of the first two categories involve the estimation of a set of 3D joint landmarks from 2D images.
By using datasets such as Human3.6m \cite{IonescuSminchisescu11} with ground truth 3D labels generated by motion capture systems, it is possible to directly regress from the input images to 3D joint positions using DNNs \cite{Pavlakos2017Coarse-to-finePose,Pavlakos2018OrdinalEstimation}.
Another option is to use the abundance of 2D ground truth labels to learn to detect 2D joint keypoints \cite{Fang2017RMPE:Estimation,Cao2017RealtimeFields}, and then ``lift" these to 3D using methods such as dictionary of skeletons \cite{Akhter2015Pose-conditionedReconstruction,Ramakrishna2012ReconstructingLandmarks,Valmadre2010DeterministicStructure} or DNNs \cite{Habibie2019InRepresentations,Zhao2016AImage,Martinez2017AEstimation}.
While the direct approach tends to overfit to indoor lab environments, the lifting strategy discards a lot of potentially useful information from the captured images.

More recently, parametric 3D human body models \cite{Loper2015SMPL:Model,Anguelov2005SCAPE:People,STAR:ECCV:2020} have been successfully used as output targets for human body shape and posture estimation.
Human body models encapsulate statistics of body shape and other priors to reduce the ambiguity of the 3D estimation problem, and can provide high-level details such as facial expressions and hand or foot articulation \cite{SMPL-X:2019}.
It is for these advantages that we choose to use the SMPL parametric human mesh model \cite{Loper2015SMPL:Model} as the 3D representation for human bodies in BodySLAM.

These parametric mesh models can be fitted end-to-end using either an optimisation-based top-down approach, a regression-based bottom-up approach, or a combination of both \cite{Bogo:ECCV:2016,kanazawaHMR18,kolotouros2019spin}.
SMPLify \cite{Bogo:ECCV:2016} proposes an optimisation-based fitting routine where the 3D mesh model is fitted to predicted 2D keypoints.
While top-down approaches such as SMPLify require a 2-step process of keypoint prediction and mesh optimisation, regression-based bottom-up approaches start directly from pixel-level information.
\cite{kanazawaHMR18} develops a direct parametric mesh regressor trained on 2D and 3D data, predicting mesh and camera parameters from an image input.
\cite{kolotouros2019spin} uses an optimisation routine in the training loop of a deep convolutional neural network to improve the supervisory signal and quality of the predictions.
In our work, we use the method of \cite{kolotouros2019spin} to initialise the human shape and posture parameters, but then refine these estimates through our joint factor graph optimisation procedure.

\subsection{Human Motion Models}

While there are many approaches to estimate human motion from videos looking only at joint locations \cite{Dabral2017LearningMotion,Hossain2017ExploitingEstimation,Pavllo20183DTraining}, our focus is on methods that use parametric human body models like SMPL.

A common approach is to enforce a smoothness prior on the posture parameters rather than define an explicit motion model.
For example, \cite{Arnab2019} does this to include temporal information in their extension of the previously introduced SMPLify routine.
By adding smoothness constraints on posture parameters and enforcing consistent body shape, the task of human shape and posture estimation from video is essentially formulated as a bundle adjustment problem.
\cite{Huang2017TowardsTime} uses multiple views of the same scene and silhouette constraints to improve single-frame fitting.
While smoothness constraints help improve estimates of the human body parameters, they are agnostic to the direction of motion and do not help to estimate scale.
As scale recovery is an important part of our work, we use a model to predict future body position and posture parameters from previous estimates, in which smoothness is implicitly included.

There are other examples that use DNNs to learn a human motion model \cite{Kanazawa2019LearningVideo,kocabas2019vibe,ling2020MVAE,rempe2021humor}.
In \cite{Kanazawa2019LearningVideo}, a network learns human kinematics by prediction of past and future image frames.
Recently, \cite{kocabas2019vibe} introduced both a temporal encoder with a gated recurrent unit (GRU) to estimate temporally consistent motion and shapes, and a training procedure with a motion discriminator that forces the network to generate feasible human motion.
While such a recurrent network is capable of generating very accurate human motion predictions, including it in a factor graph optimisation would be difficult as each hidden state would need to become a state variable.
Instead, we opt to use a multi-layer perceptron that takes the body shape parameters and the past $n$ body posture estimates as input to predict the current posture parameters.
This formulation allows us to greatly simplify the factor graph and will allow for much easier marginalisation of past states as part of future work.

\subsection{Visual SLAM with Dynamic Objects}

Most visual SLAM systems assume that the scene they are mapping and tracking against is static (e.g.\ ORB-SLAM \cite{Mur-Artal2017ORB-SLAM2:Cameras}).
Many of these systems are still capable of handling some small dynamic regions of the scene as any keypoints in these areas will be treated as outliers when using robust methods; however, they will fail if the dynamic objects occupy large segments of the frame as is often the case in image sequences capturing human motion.
Some systems are designed to handle dynamic environments through the use of elimination strategies, where moving objects are detected and removed from the tracking and mapping processes (e.g.~\cite{Scona2018StaticFusion:Environments,Bescos2018DynaSLAM:Scenes,jaimezICRA2017fastodom,barnesICRA2018distractor,daiTPAMI2020rgbdslam,jiICRA2021realtime}).
While this strategy offers an improvement over the static scene assumption in standard SLAM systems, it still requires that enough static background is visible for robust tracking and throws away any information contained in the dynamic parts of the environment.
Alternative methods explicitly model motion with techniques such as rigid object-level maps \cite{Xu2019MID-Fusion:SLAM,runzICRA2017cofusion,Runz2018MaskFusion:Objects,barsanICRA2018robustdense}.

More recently, there have been some approaches that estimate the camera motion and dynamic scenes simultaneously, without any \textit{a priori} knowledge of the environment (e.g.~\cite{heneinICRA2018exploitingrigid,juddIROS2018mvo,heneinICRA2020dynamicslam}).
While these systems are limited to estimating rigid-body motion, AirDOS \cite{qiuICRA2022airdos}, the work most closely related to our own, extends this line of work to include articulated objects such as humans.
AirDOS uses a combination of rigidity and motion constraints to create articulated models for humans and vehicles within a factor graph framework.
By jointly optimising for the object motion, object 3D structure and the camera trajectory, improved camera tracking over an elimination strategy baseline is shown.
While AirDOS presents impressive results, we are able to make a number of key improvements over their approach.
Firstly, whereas AirDOS depends on stereo input, BodySLAM works with only a monocular camera, making it suitable to more ``in the wild" applications.
Unlike stereo SLAM systems, standard monocular SLAM systems are unable to observe the scale of the scene; however, we demonstrate that by simultaneously optimising for the human posture, shape and centre poses, BodySLAM is able to recover the metric scale.
Secondly, since the focus of AirDOS is on improved camera tracking and not human motion estimation, it uses a simple skeleton-based model of human rigidity.
It furthermore does not use a a learned temporal model, but uses motion constraints for the rigid body segments.
BodySLAM, on the other hand, uses the SMPL parametric mesh model, allowing for full human body reconstructions, and uses a temporal human motion model to predict translation, orientation, and posture change.
Finally, while AirDOS provides quantitative camera tracking results, only a few qualitative results for the object tracking and structure estimation are provided.
In our experiments on real-world datasets, we quantitatively demonstrate that BodySLAM improves not only camera tracking, but the estimation of the human body centre poses, shape and posture parameters.

\section{Methodology}
\subsection{Preliminaries and Notation}
We use the following notation throughout this work:
a reference frame is denoted as $\cframe{A}$, and vectors expressed in it are written as $\pos{A}{}$ or translations $\pos{A}{OP}$ with $O$ and $P$ as start and endpoints, respectively.
The homogeneous transformation from reference frame $\cframe{B}$ to $\cframe{A}$ is denoted as $\T{A}{B}$.
Its rotation matrix is denoted as $\C{A}{B}$, whose minimal representation we parameterise as an axis-angle rotation vector in the respective frame, $\rotvec{A}{AB}$.
Commonly used reference frames are the static world frame $\cframe{W}$, the camera centric frame $\cframe{C}$, and the human body centric frame $\cframe{H}$.
Camera frames are indexed by time steps $k$, and body joints by $j$.
Measurements of quantity $\mbf{z}$ are denoted with a tilde, $\mbftilde{z}$. 

\subsection{System Overview}
\begin{figure}
    \centering
    \captionsetup{width=\textwidth}
    \includegraphics[width=0.95\textwidth]{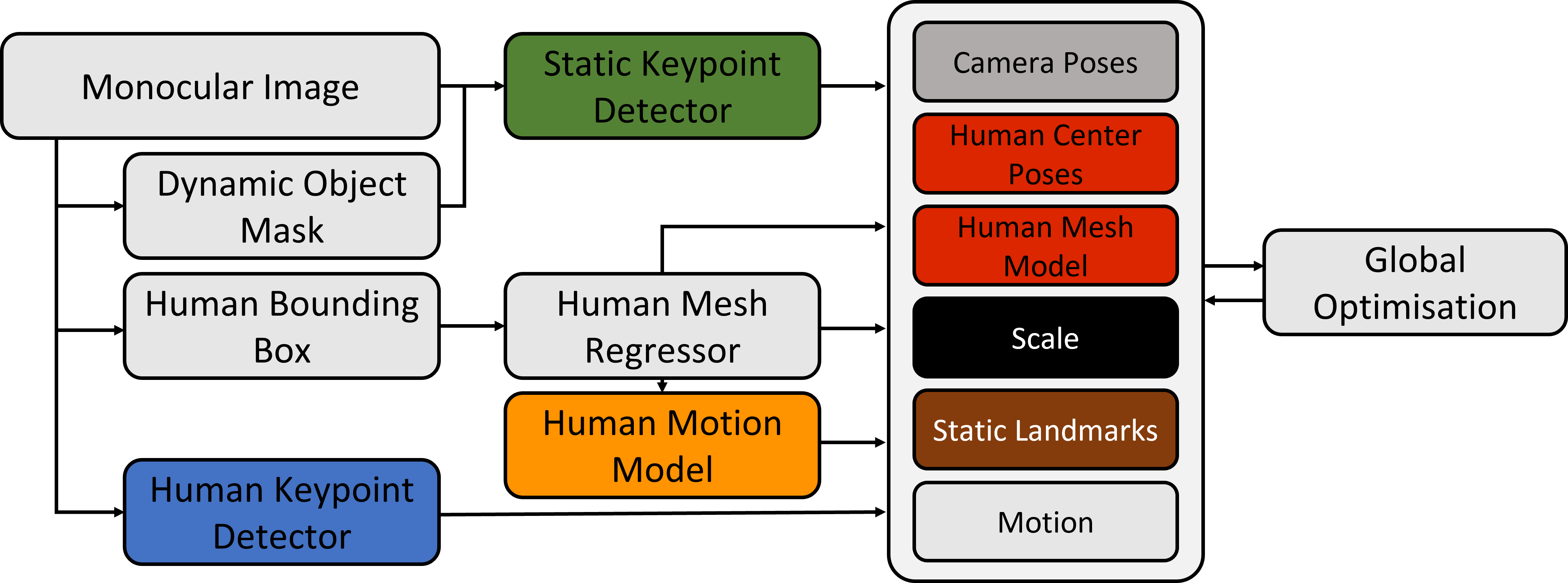}
    \caption{The BodySLAM framework.}
    \label{fig:pipeline}
\end{figure}
An overview of the BodySLAM framework is provided in Fig.~\ref{fig:pipeline}.
BodySLAM can be split into two main components: a front-end that performs the pre-processing of the input image stream, and the back-end that performs the graph optimisation (bundle adjustment).

In the front-end, BRISK keypoints are extracted from each image for use in camera tracking \cite{Leutenegger2011BRISK:Keypoints}.
The monocular bundle adjustment problem is then initialised with static landmarks and camera poses obtained by applying RANSAC on the keypoints.
Each image is also processed by an instance segmentation network \cite{wu2019detectron2} to detect human bounding boxes, and by OpenPose \cite{Cao2017RealtimeFields} to extract human keypoints.
From these, human mesh parameters are estimated using a Human Mesh Regressor (HMR) network \cite{kolotouros2019spin}.
Finally, human motion and posture estimates are obtained from our deep motion model (see Section \ref{sec:hmm}).

Many state-of-the-art methods perform bundle adjustment only on keyframes, a selected subset of the total frames; however, to avoid complications to the training and inference of the motion model by having a variable frame rate, we assume a constant frame rate of 30 FPS and estimate the camera pose, human centre pose, and human shape and posture parameters at every frame.

\subsection{3D Representation of the Human Body}
The SMPL model \cite{Loper2015SMPL:Model} is a parametric human mesh model that supplies a function, $\mathcal{M}(\mbs{\beta}, \mbs{\theta})$, of the shape parameters, $\mbs{\beta}$, and posture parameters, $\mbs{\theta}$, and returns a mesh, $_H\mbf{M} \in \mathbb{R}^{N \times 3}$, as a set of $N = 6890$ vertices in the human-centric frame $\cframe{H}$.
The body shape is parameterised as a 10-dimensional vector, $\mbs{\beta}\in \mathbb{R}^{10}$, where the components are determined by a principal component analysis to capture as much of the variability in human body shape as possible.
The body posture parameters, $\mbs{\theta}$, represent each of the 23 joint orientations in the respective parent frame in axis-angle representation with dimensionality $\mathrm{dim}(\mbs{\theta}) = 23 \times 3 = 69$.
With this kinematic tree representation, it is easy to encode the same posture for different human shapes, as the joint positions can be defined as a linear combination of a set of mesh vertices.
For $J$ joints, a linear regressor $\mbf{W} \in \mathbb{R}^{J \times N}$ is pre-trained to regress the $J$ 3D body joints as $\myvec{X}{H} = \mbf{W} \: _H\mbf{M}$ with $\myvec{X}{H} \in \mathbb{R}^{J \times 3}$.

The 6D human centre pose is defined as $\mbf{x}^\mathrm{human} = [\pos{W}{WH}, \rotvec{W}{WH}] \in \real^6$, where $\pos{W}{WH}$ is the body centre position with respect to the static world frame, and $\rotvec{W}{WH}$ is the body centre orientation represented as a rotation vector.

\subsection{Measurements of Human Body Parameters}
\label{method:relative_pose_prediction}
\subsubsection{SMPL Parameter Prediction}
The state-of-the-art method from Kolotouros \et \cite{kolotouros2019spin} is used to generate measurements of the SMPL parameters $\{\mbstilde{\beta}, \mbstilde{\theta}\}$ in each frame.
This method uses an HMR network to regress the SMPL parameters from a normalised and cropped image of a human.
The cropped human image is obtained by detecting human bounding boxes using Detectron2 \cite{wu2019detectron2}.

\subsubsection{Human Centre Pose Prediction}
We also use the network in \cite{kolotouros2019spin} to generate measurements of the 6D transformation between the camera and human centre frames.
In addition to the SMPL parameters, the HMR network also predicts a weak perspective camera model.
As detailed in \cite{hpe3d}, after rectifying and removing focal distortion from the images of a calibrated camera, we can use the weak perspective camera parameters and the rotation between the camera centre and the centre of the detected bounding box to derive the transformation $\mbfhtilde{T}_{CH}$.

\subsection{Human Motion Model} \label{sec:hmm}
\subsubsection{Formulation}
For the motion model, we use a deep neural network.
Its design follows the architecture from Martinez \et \cite{Martinez2017AEstimation}, but instead of a recurrent unit used for the temporal encoding, we use a sequential backbone only consisting of a multi-layer perceptron.
The motion model uses the past $n$ body posture estimates $\mbs{\Theta} = [\mbs{\theta}_{k\text{-}n}, \dots, \mbs{\theta}_{k\text{-}1}]$ and the (constant) body shape $\mbs{\beta}$ as inputs, to predict the current body posture, $\mbs{\theta}_{k}^{\mathrm{pred}}$, and relative human centre translation, $\pos{H_{k\text{-}1}}{H_k H_{k\text{-}1}}^{\mathrm{pred}}$.
This prediction is used as a motion measurement in the optimisation problem discussed in Section \ref{method:optim}.
The model is trained using the loss:
\begin{equation}
\label{eq:loss}
    \mathcal{L} = {c}_{p} + \lambda {c}_{\theta},
\end{equation}
where ${c}_{p} = \|\pos{H_{k\text{-}1}}{H_{k\text{-}1} H_k}^{\mathrm{true}} - \pos{H_{k\text{-}1}}{H_{k\text{-}1} H_k }^{\mathrm{pred}} \|^2$ is the error on the predicted relative position, and ${c}_{\theta} = \|\mbs{\theta}_{k}^{\mathrm{true}} - \mbs{\theta}_{k}^{\mathrm{pred}} \|^2$ is the error on the human body posture.
The global orientation change of the human centre frame is predicted as part of the 24 body joint angles.
The relative weighting between the two error terms $\lambda$ is set to $1/24$, weighting the position error as much as each rotation in the kinematic tree.
Since the input and output values of the network were normalised with the respective means and standard deviations of the dataset, the loss values quickly converge.
No robust cost function was used for the optimisation loss.

\subsubsection{Architecture}
For the human posture and shape encoder, we use a 3-layer linear encoder, with an input layer of size $(n \times 3 \times 24) + 10$ for the $n$ previous human body posture and shape estimates, and two fully connected layers of size 1024.
We use 2 separate decoders for the human position change and human body posture, each with two fully connected layers, and a final layer of size 3 and 72, respectively.
The human posture decoder has an additional residual connection to the most recent posture $\mbs{\theta}_{k} = \mbs{\theta}_{k\text{-}1} + \mbs{\theta}^{\mathrm{pred}}$.

\subsubsection{Training Procedure}
The human motion model is trained on a subset of the AMASS dataset \cite{AMASS:2019}, only considering sequences with walking, running, or jogging motions.
Since we predict the pose change between two consecutive camera frames, we processed the data of the AMASS dataset to match the most common camera frame rate of 30 FPS.
The network uses a fixed number of input frames for its prediction.
To avoid overlapping sequences, we processed the dataset into chunks containing the required number of input frames and a target frame.
We experimented with different sequence lengths of $n = \{2, 4, 8, 16\}$, but did not notice a drastic change in performance with longer sequences, and so trained the model with $n = 2$.
This left us with more than 340,000 sequences, from which we reserved 20\% for validation.
We trained the network until convergence (100 epochs), with a batch size of 4096 on a single Nvidia GeForce GTX1080Ti GPU.
We used the Adam optimiser \cite{Kingma2015Adam:Optimization} with a learning rate of $1 \times 10^{-3}$.

\subsection{Factor Graph Formulation}
\label{method:optim}
\subsubsection{Overview}
During bundle adjustment, we jointly optimise camera poses, $\mbf{x}^\mathrm{cam} = [{_W}\mbf{r'}_{WC_k}, \rotvec{W}{WC_k}]$, human centre poses, $\mbf{x}^\mathrm{human} = [\pos{W}{WH_k}, \rotvec{W}{WH_k}]$, body shape, $\mbs{\beta}$, postures, $\mbs{\theta}_{k}$, and $L$ static landmarks, $\lmh{W}_l$ with $l\in{1,\ldots,L}$. For faster convergence, we over-parameterise the camera position as $\pos{W}{WC_k} = s\ {_W}\mbf{r'}_{WC_k}$, where the scale $s$ is estimated along with the (scaled) position vector $\mbf{r'}_{WC_k}$. Thus we estimate $\mbf x = [s, \mbs \beta, \mbf{x}^\mathrm{cam}_1, \mbf{x}^\mathrm{human}_1, \mbs \theta_1, \ldots , \mbf{x}^\mathrm{cam}_K, \mbf{x}^\mathrm{human}_K, \mbs \theta_K, \lmh{W}_1, \ldots, \lmh{W}_L]$.

\begin{figure}[t]
    \centering
    \captionsetup{width=\textwidth}
    \includegraphics[width=0.85\textwidth]{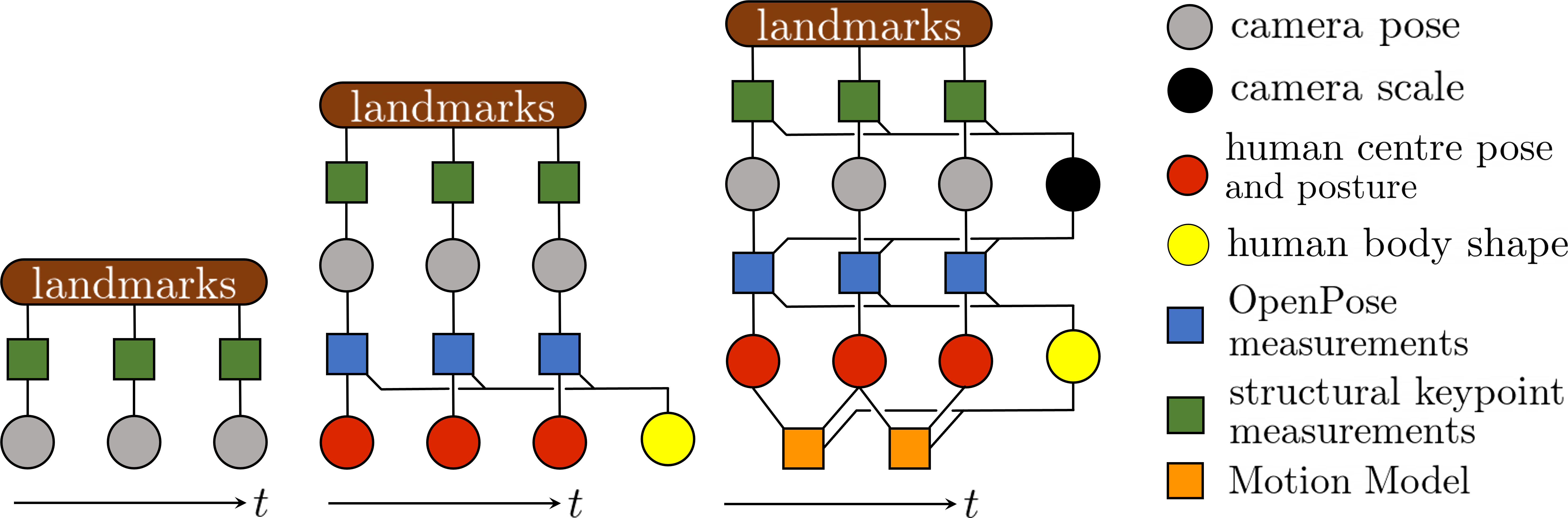}
    \caption{Baseline and proposed Factor Graphs of the visual SLAM problem. Without human bodies, the monocular SLAM Factor Graph has no relation between successive camera poses (left). A naive formulation of SLAM problem with human 2D keypoint measurements as dynamic keypoints fails to create a relation between successive frames (middle). Our proposed BodySLAM formulation adds a motion model that predicts human centre pose and posture, and creates a temporal constraint for camera and human center poses (right). Additionally, we overparameterise the camera pose with an additional scale factor $s$, to achieve faster convergence of our Bundle Adjustment.}
    \label{fig:factor_graph}
\end{figure}

Fig.~\ref{fig:factor_graph} shows the respective factor graphs for three different SLAM formulations.
The variables to be estimated are drawn as circles and the measurements are drawn as boxes.
The measurements include 3D landmarks, OpenPose keypoint measurements and relative motion factors.
The leftmost factor graph describes the traditional monocular SLAM problem without temporal constraints between the camera poses.
The middle factor graph shows the na\"{i}ve formulation, where human reprojection error terms are included in blue, but there are still no temporal constraints.
The rightmost factor graph includes measurements from our learned motion model, creating temporal constraints between the estimated camera and human centre poses, rendering scale observable.
Through our experimental results, we will demonstrate that using the rightmost factor graph leads to the most accurate estimates of both human body parameters and camera poses.
Therefore, the total cost function optimised by BodySLAM is given by:
\begin{align}
    c(\mbf{x}) &= \lambda^\mathrm{lm}\sum_{k=1}^{K} \sum_{l=1}^{L} {\mbf{e}_{k,l}^{\mathrm{lm}}}^{\mathrm{T}} \mbf{e}_{k,l}^{\mathrm{lm}} 
    + \sum_{k=1}^{K} \sum_{j=1}^{J} \lambda_{k,j}^\mathrm{joints}{\mbf{e}_{k,j}^{\mathrm{joints}}}^{\mathrm{T}} \mbf{e}_{k,j}^{\mathrm{joints}}
    + \lambda^{\mathrm{mm}} \sum_{k=2}^{K} {\mbf{e}_{k}^{\mathrm{mm}}}^{\mathrm{T}} \mbf{e}_{k}^{\mathrm{mm}} \nonumber \\
    &+ \lambda^{\mathrm{posture}} \sum_{k=1}^{K} {\mbf{e}_{k}^{\mathrm{posture}}}^{\mathrm{T}} \mbf{e}_{k}^{\mathrm{posture}} \nonumber + \lambda^{\mathrm{shape}} \sum_{k=1}^{K} {\mbf{e}^{\mathrm{shape}}_k}^{\mathrm{T}}\mbf{e}^{\mathrm{shape}}_k ,
\end{align}
\noindent
where $\mbf{e}_{k,l}^{\mathrm{lm}}$, $\mbf{e}_{k,j}^{\mathrm{joints}}$, $\mbf{e}_{k}^{\mathrm{mm}}$, $\mbf{e}_{k}^{\mathrm{posture}}$, and $\mbf{e}^{\mathrm{shape}}$ are the error terms for the structural landmarks, human joint landmarks, human motion, body posture, and body shape, respectively.
Each of these error terms is discussed in detail below.
The structural landmarks are weighted by $\lambda^{\mathrm{lm}} = 64/b^2$, where $b$ is the size of the BRISK keypoint ($b = 2.25\mathrm{px}$ in our implementation).
The human joint landmarks are weighted by $\lambda_{k,j}^{\mathrm{joints}} = \sigma_{k,j}^{-2}$, where $\sigma_{k,j}$ is confidence predicted by OpenPose.
Finally, in our implementation, we use $\lambda^{\mathrm{mm}} = 100$, $\lambda^{\mathrm{posture}} = 1$, and $\lambda^{\mathrm{shape}} = 1$, the values of which were determined experimentally.

\subsubsection{Reprojection Error for Structural Landmarks}
For the structural landmarks, $\lmh{W}_l$, we use the standard reprojection error term (as described in \cite{Leutenegger2015Keyframe-basedOptimization}):
\begin{equation}
    \mbf{e}_{k,l}^{\mathrm{lm}} = \mbftilde{z}_{k,l} - \mbf{u}(\T{W}{C_k}^{-1} \lmh{W}_{l}),
\end{equation}
\noindent
where $\mbf{u}(\cdot)$ denotes the camera projection model and $\mbftilde{z}_{k,l}$ is the 2D keypoint measurement of the static landmark $\lmh{W}_l$ in frame $k$.
The measurements are BRISK keypoints filtered for outliers by 3D-2D RANSAC.
To handle any remaining outliers, we use the robust Cauchy loss function on the error term.

\subsubsection{Reprojection Error for Human Joint Landmarks}
As done in \cite{Bogo:ECCV:2016,kolotouros2019spin}, we use OpenPose keypoints and their corresponding confidence values as measurements in the joint reprojection error term, employing the same standard reprojection error term as for the structural keypoints:
\begin{equation}
    \mbf{e}_{k,j}^{\mathrm{joints}} = \mbftilde{z}_{k,j} - \mbf{u}(\T{W}{C_k}^{-1}\T{W}{H_k} \lmh{H}_{j,k}(\mbs \theta_k, \mbs \beta)),
\end{equation}
\noindent
where $\mbftilde{z}_{k,j}$ denotes the OpenPose keypoint measurement, and $\lmh{B}_{j,k}(\mbs \theta_k, \mbs \beta)$ denotes the SMPL joint $j$ in frame $k$ expressed as homogeneous points in the human centered frame $\cframe{H}$ as a function of the posture $\mbs \theta_k$ and shape $\mbs \beta$.
To increase robustness in particular with (self-)occluded joints and noisy keypoint detections, we use the Geman-McLure robust cost function in line with \cite{Bogo:ECCV:2016}.

Note that the SMPL parameters $\{\mbs{\theta}, \mbs{\beta}\}$ can be directly initialised from the predictions as explained in Section \ref{method:relative_pose_prediction}. In turn, the human poses can be initialised as $\T{W}{H_k} = \T{W}{C_k}\mbfhtilde{T}_{C_kH_k}$.

\subsubsection{Human Motion Error}
Our motion model is used to predict the expected human translation in the next camera frame $\postilde{H_{k\text{-}1}}{H_{k\text{-}1} H_k} := \pos{H_{k\text{-}1}}{H_{k\text{-}1} H_k }^{\mathrm{pred}}$.
This prediction is used in the human motion error term:
\begin{equation}
    \mbf{e}_{k}^{\mathrm{mm}} = \postilde{H_{k\text{-}1}}{H_{k\text{-}1} H_k} - 
    \C{W}{H_{k\text{-}1}}^T  \left( \pos{W}{W H_k} - \pos{W}{W H_{k\text{-}1}} \right),
\end{equation}
\noindent
where $\C{W}{H_{k\text{-}1}}$ denotes orientation of the human pose.
One can notice the similarity of the human motion error term to the position error term in Eq.~(\ref{eq:loss}) used during the training of the human motion model.

\subsubsection{Human Body Posture Error}
Our motion model predicts the expected human body posture in the current frame, $\mbstilde{\theta}_{k}$.
The posture error is formulated as:
\begin{equation}
    \mbf{e}_{k}^{\mathrm{posture}} = \mbstilde{\theta}_{k} - \mbs{\theta}_{k}.
\end{equation}

\subsubsection{Human Shape Error}
In each frame, we use an HMR network to predict the SMPL parameters $\{\mbstilde{\beta}_k, \mbstilde{\theta}_k\}$.
As $\mbs{\beta}$ should be constant over the image sequence, we constrain this to be a constant using the body shape error term:
\begin{equation}
    \mbf{e}_{k}^{\mathrm{shape}} = \mbstilde{\beta}_{k} - \mbs{\beta}.
\end{equation}

\subsubsection{Optimisation Procedure and Implementation Details}

To initialise the full BodySLAM optimisation, we perform a straightforward implementation of bundle adjustment using the BRISK keypoints, RANSAC-based pose initialisation and optimisation using the Google Ceres optimiser \cite{ceres} to obtain estimates for the camera poses and landmark locations up to an arbitrary scale.
After this initialisation, we use a first-order nonlinear least squares optimisation in PyTorch \cite{Paszke2017AutomaticPyTorch} for bundle adjustment.
However, to increase convergence speed and performance, one could easily implement a second-order optimisation.

Since the first-order optimisation is prone to converge at local minima, we use a two-step optimisation schedule often used in human mesh optimisation \cite{Arnab2019,Bogo:ECCV:2016,kolotouros2019spin}.
First, we optimise only the relative transformations between the cameras and the human body centres, the transformations between the cameras and the world frame, and the static landmarks, while keeping all other variables constant.
The purpose of this optimisation step is to roughly estimate the scale of the given scene, and the camera and human centre trajectories.
We optimise until convergence, finding approximately 200 iterations to be sufficient.
In a second step, we optimise all variables for another 100 iterations after which we always reached convergence in our experiments.

\section{Experimental Results}
To evaluate BodySLAM, we collected a dataset of 10 video sequences of human motion captured by a moving monocular camera.
A motion capture system was used to collect ground truth camera trajectories and human body parameters.
This dataset is described in more detail in Section \ref{sec:res:dataset}.

For evaluation, we consider the standard metric of average trajectory error for the camera trajectory (C-ATE), the human centre trajectory (H-ATE), and the human joint trajectories (J-ATE).
Furthermore, we align the estimated camera trajectories in $\mathrm{Sim}(3)$, and interpret the aligned scale $s_{\mathrm{Sim(3)}}$ as scale error.
With a perfect scale estimation, this estimated scale would be $1.0$.

As one of the main contributions of BodySLAM is its ability to estimate metric scale from a monocular image sequence, we first evaluate how well it can recover the scale of a scene when it is given the scale-perturbed ground truth camera trajectory.
The results of this preliminary study are provided in Section \ref{sec:res:scale}.
In Section \ref{sec:res:joint}, we evaluate the ability of the full BodySLAM system to estimate the camera poses, human body centre poses, and human joint positions for all sequences in our dataset.
Finally, in Section \ref{sec:res:mm}, we perform an ablation study on our motion model, showing that our model is able to provide meaningful estimates of human motion and improves estimates of the trajectories.

\subsection{Custom Human Pose Estimation Dataset} \label{sec:res:dataset}
Current real-world evaluation datasets are lacking at least one of the following properties: a moving camera, known camera intrinsics, ground truth camera motion, a moving human subject, or ground truth human body parameters.
An extensive but incomplete list of these can be found in Table~\ref{tab:dataset_overview} in the Appendix.
Due to these limitations, we perform the evaluation on our own custom dataset that captures the ground truth camera poses of a moving camera with known intrinsics, along with human centre poses and posture parameters for moving humans.
Ground truth values were obtained using a Vicon motion capture system.
The dataset consists of 10 separate sequences, each between 600 to 1000 frames, totalling 8174 images.
The dataset is split into easy, medium, and hard sequences, categorised by the speed and complexity of the camera motion.

\subsection{Study of Scale Recovery Through Motion Factors} \label{sec:res:scale}
To evaluate the ability of BodySLAM to recover metric scale, we perform an ablation study where structural landmark measurements are ignored and where we initialise the camera trajectory with the Vicon ground truth poses after scaling the camera positions with an arbitrary perturbation factor between $0.1$ and $5.0$.
We then freeze the $\mathrm{Sim}(3)$ transformations of the camera poses in the graph, only allowing the scale of the camera trajectory, $s$, to be optimised along with the human body parameters.
As shown in Fig.~\ref{fig:scale_recovery}, after initial perturbation by $1 / s'$, we are able to reduce the H-ATE to approximately the value it had been before perturbation.
Furthermore, we show that the scale is correctly estimated with a maximum error of 26.5\% for the most extreme perturbation factor of $5.0$ (as found in Table~\ref{tab:scale_recovery} in the Appendix).
For smaller initial perturbations, the estimated scale deviates by only roughly 5\% from the true scale.

\begin{table}[t]
\begin{center}
\captionsetup{width=\textwidth}
\caption{Comparison of absolute trajectory errors: camera (C-ATE), human (H-ATE), and joints (J-ATE) plus camera trajectory scale on test sequences for the baseline without motion model factor (``no MM'') and the full BodySLAM (ours). Position errors are computed after SE(3) alignment to account for the difference in the Vicon global frame and visual world frame $\cframe{W}$. We also aligned the scale-unaware baseline using Sim(3), which reveals that our method still outperforms it regarding human centre position and joint locations, and performs on-par regarding camera pose accuracy. Note, however, that \textbf{in the wild Sim(3) alignment is not possible}, since it requires access to the ground truth camera trajectory -- illustrating the usefulness of our method.}
\label{tab:ate_and_scale_comp}
\resizebox{\textwidth}{!}{
\begin{tabular}{l|rrrr|rrrr|rrrr|rr}
\textbf{}     & \multicolumn{4}{c|}{\textbf{C-ATE {[}mm{]}}}                                                                              & \multicolumn{4}{c|}{\textbf{H-ATE {[}mm{]}}}                                                                              & \multicolumn{4}{c|}{\textbf{J-ATE {[}mm{]}}}                                                                              & \multicolumn{2}{c}{\multirow{2}{*}{\textbf{Scale Error}}} \\
\textbf{}     & \multicolumn{2}{c|}{SE(3)}                                      & \multicolumn{2}{c|}{Sim(3)}                             & \multicolumn{2}{c|}{SE(3)}                                      & \multicolumn{2}{c|}{Sim(3)}                             & \multicolumn{2}{c|}{SE(3)}                                      & \multicolumn{2}{c|}{Sim(3)}                             & \multicolumn{2}{c}{}                                      \\ \hline
\textbf{Seq.} & \multicolumn{1}{c}{no MM} & \multicolumn{1}{c|}{(ours)}         & \multicolumn{1}{c}{no MM} & \multicolumn{1}{c|}{(ours)} & \multicolumn{1}{c}{no MM} & \multicolumn{1}{c|}{(ours)}         & \multicolumn{1}{c}{no MM} & \multicolumn{1}{c|}{(ours)} & \multicolumn{1}{c}{no MM} & \multicolumn{1}{c|}{(ours)}         & \multicolumn{1}{c}{no MM} & \multicolumn{1}{c|}{(ours)} & \multicolumn{1}{c}{no MM}   & \multicolumn{1}{c}{(ours)}  \\ \hline
E 1           & 108.1                     & \multicolumn{1}{r|}{\textbf{44.3}}  & 42.4                      & 43.5                        & 114.0                     & \multicolumn{1}{r|}{\textbf{56.3}}  & 106.5                     & 56.0                        & 154.8                     & \multicolumn{1}{r|}{\textbf{118.8}} & 149.8                     & 115.9                       & 0.671                       & \textbf{0.940}              \\
E 2           & 193.0                     & \multicolumn{1}{r|}{\textbf{50.8}}  & 50.7                      & 49.1                        & 178.6                     & \multicolumn{1}{r|}{\textbf{55.0}}  & 182.8                     & 53.0                        & 210.9                     & \multicolumn{1}{r|}{\textbf{124.2}} & 214.6                     & 120.5                       & 0.581                       & \textbf{1.030}              \\
E 3           & 122.3                     & \multicolumn{1}{r|}{\textbf{40.9}}  & 40.8                      & 38.9                        & 220.8                     & \multicolumn{1}{r|}{\textbf{160.9}} & 217.1                     & 160.4                       & 247.7                     & \multicolumn{1}{r|}{\textbf{196.1}} & 244.3                     & 192.8                       & 0.738                       & \textbf{0.981}              \\
M 1           & 314.7                     & \multicolumn{1}{r|}{\textbf{92.6}}  & 93.3                      & 91.0                        & 326.9                     & \multicolumn{1}{r|}{\textbf{201.4}} & 314.6                     & 200.0                       & 368.8                     & \multicolumn{1}{r|}{\textbf{236.2}} & 355.3                     & 236.2                       & 0.502                       & \textbf{0.978}              \\
M 2           & 241.6                     & \multicolumn{1}{r|}{\textbf{64.0}}  & 57.9                      & 61.1                        & 368.1                     & \multicolumn{1}{r|}{\textbf{295.4}} & 353.5                     & 297.1                       & 408.6                     & \multicolumn{1}{r|}{\textbf{323.3}} & 392.0                     & 325.4                       & 0.507                       & \textbf{0.936}              \\
M 3           & 392.1                     & \multicolumn{1}{r|}{\textbf{128.3}} & 128                       & 128.1                       & 311.2                     & \multicolumn{1}{r|}{\textbf{121.7}} & 305.9                     & 122.2                       & 335.5                     & \multicolumn{1}{r|}{\textbf{170.2}} & 326.4                     & 171.9                       & 0.354                       & \textbf{0.985}              \\
M 4           & 203.1                     & \multicolumn{1}{r|}{\textbf{92.4}}  & 91.3                      & 90.9                        & 153.7                     & \multicolumn{1}{r|}{\textbf{70.4}}  & 142.8                     & 71.7                        & 184.2                     & \multicolumn{1}{r|}{\textbf{123.0}} & 175.0                     & 125.3                       & 0.635                       & \textbf{0.951}              \\
D 1           & 310.4                     & \multicolumn{1}{r|}{\textbf{108.1}} & 107                       & 103.5                       & 303.7                     & \multicolumn{1}{r|}{\textbf{136.7}} & 293.9                     & 135.9                       & 330.6                     & \multicolumn{1}{r|}{\textbf{180.6}} & 321.1                     & 181.1                       & 0.445                       & \textbf{0.951}              \\
D 2           & 316.0                     & \multicolumn{1}{r|}{\textbf{100.0}} & 98.9                      & 95.2                        & 297.6                     & \multicolumn{1}{r|}{\textbf{91.2}}  & 283.5                     & 90.6                        & 320.6                     & \multicolumn{1}{r|}{\textbf{142.6}} & 306.8                     & 143.8                       & 0.430                       & \textbf{0.933}              \\
D 3           & 246.9                     & \multicolumn{1}{r|}{\textbf{115.3}} & 113                       & 114.0                       & 235.4                     & \multicolumn{1}{r|}{\textbf{149.9}} & 228.8                     & 149.5                       & 264.5                     & \multicolumn{1}{r|}{\textbf{183.9}} & 258.0                     & 185.0                       & 0.494                       & \textbf{0.942}             
\end{tabular}
}

\end{center}
\end{table}

\subsection{BodySLAM: Optimisation of Camera Poses and Human States} \label{sec:res:joint}
In Table~\ref{tab:ate_and_scale_comp}, we present the main results for our full BodySLAM system.
As a baseline for comparison, we apply classic monocular bundle adjustment with na\"{i}ve human motion tracking (i.e. only considering the unary OpenPose measurements and no motion model).
This baseline, represented by the the middle factor graph in Fig.~\ref{fig:factor_graph}, also does not use the scale overparameterisation of BodySLAM.

Our results show that both the C-ATE and the H-ATE, after $SE(3)$ alignment, are significantly lower than the baseline method.
The average improvement through the introduction of the motion model is 48.3\% for the root trajectory, and 35.6\% for all joints combined.

Furthermore, we show that the motion model factor is able to accurately recover the scale, with scale estimation errors in the range of $0.93$ to $1.03$, compared to errors between $0.35$ to $0.74$ for the baseline, as shown in Table \ref{tab:ate_and_scale_comp}.
However, one must note that the values of the scale error are completely arbitrary, since they depend only on the random initialisation of the camera poses.
The C-ATE with $SE(3)$ alignment for BodySLAM shows a huge improvement compared to the baseline results.
This improvement is attributed to the correctly estimated scale, as the results after $\mathrm{Sim}(3)$ alignment clearly show.
After $\mathrm{Sim}(3)$ alignment, the C-ATEs differ only by a few millimeters, which can be caused by small convergence differences during the first order optimisation.

We performed an additional experiment, ablating the necessity of the structural landmarks in the joint optimisation.
For this, we performed an optimisation of the camera poses and human states, with and without structural landmarks to constrain the camera poses.
The initial camera poses were randomly perturbed ($\pm$100 mm, $\pm$0.01 rad) and optimised until convergence.
The optimisation reduced the mean camera position error from 228.96 mm after perturbation to 155.5 mm with landmarks, and 158.6 mm without landmarks (1.94\% error increase), highlighting BodySLAM's versatility.
The complete results are found in Table~\ref{tab:struct_pert} in the Appendix.

\subsection{Motion Model Prediction Performance} \label{sec:res:mm}
To evaluate the contribution of our motion model, we analyse the performance of the model on the test split of our preprocessed AMASS dataset.

\begin{figure}
\centering
\begin{minipage}{.5\textwidth}
  \centering
  \captionsetup{width=0.9\linewidth}
  \includegraphics[width=0.95\linewidth]{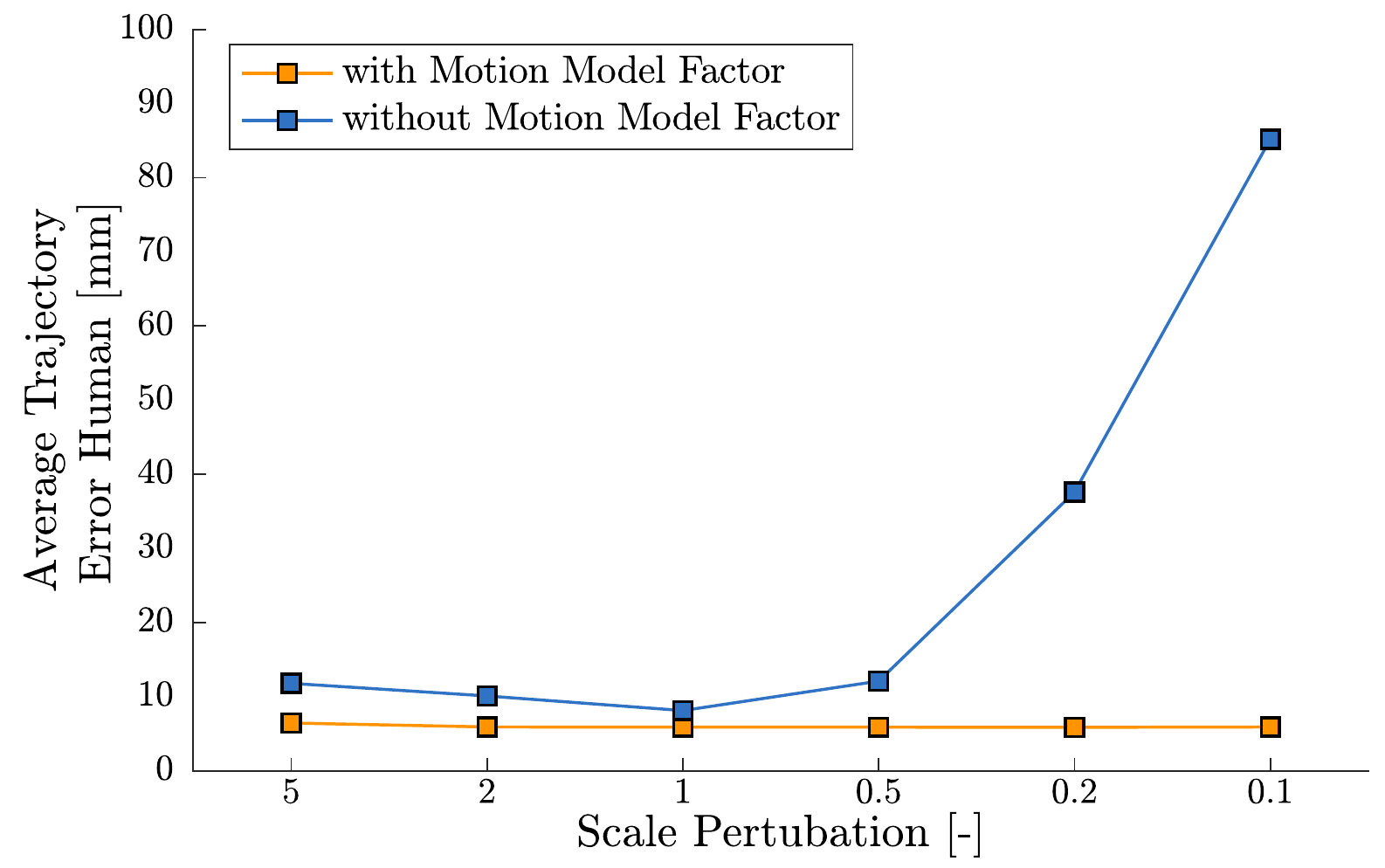}
  \captionof{figure}{Average Human Trajectory Error, $SE(3)$ aligned. Without Motion Model, the scale is not recovered.}
  \label{fig:scale_recovery}
\end{minipage}%
\begin{minipage}{.5\textwidth}
  \centering
  \captionsetup{width=0.9\linewidth}
  \includegraphics[width=0.95\linewidth]{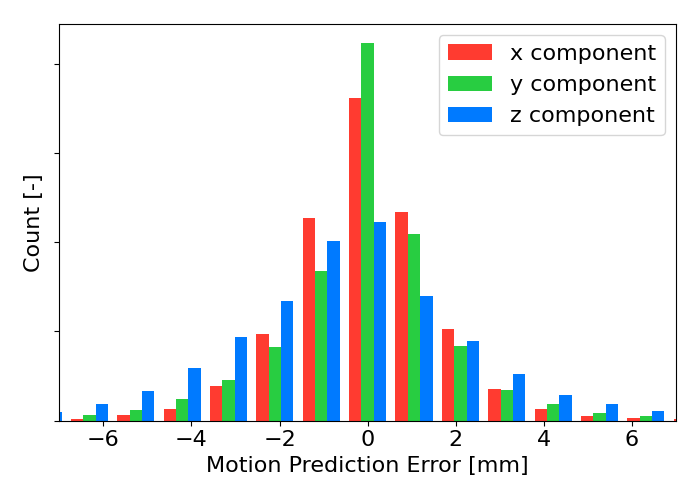}
  \captionof{figure}{Human motion prediction error histogram in x, y, and z direction.
  }
  \label{fig:hist_error}
\end{minipage}
\end{figure}

The mean translation prediction error is 3.4mm for a mean per-frame position change of 18.4mm.
The distribution of the error vector components is shown in Figure \ref{fig:hist_error}.
Please note that in the SMPL body model, the human body coordinate frame $\cframe{H}$ is aligned to the sagittal axis (forward) in the $z$-direction, and the longitudinal axis (upward) in $y$-direction.
Despite a relatively large standard deviation, our prediction errors are centered around zero, and, as previous results have shown, can significantly improve the estimation of the scale and the human centre and joint trajectories when included in the graph optimisation.

\section{Conclusions and Future Work}
We have presented BodySLAM, a novel monocular SLAM system that jointly optimises a set of 6D camera poses, 3D landmarks, human body centre poses, and human body shape and posture parameters in a factor graph formulation.
To the best of our knowledge, this is the first system to jointly estimate for the configuration of a dense human body mesh model and the trajectory of a moving monocular camera.
We have also introduced a human motion model to constrain sequential body postures and to help observe scale.
To validate our contribution, we have collected a series of video sequences of human motion captured by a moving monocular camera, with ground truth camera trajectories and human body parameters from a motion capture system.
Through a number of experiments, we have demonstrated that our joint formulation consistently results in improvements to both camera tracking and human body estimation.

In future work, we plan to extend the BodySLAM method into an incremental and real-time SLAM system.
We would also like to explore more complex architectures for the human motion model and evaluate whether or not any gains in accuracy offset the increased complexity of the factor graph formulation that these architectures would require.
Finally, one limitation of our method is that it assumes the intrinsic parameters of the camera are given, which may not always be the case.
Some very recent work \cite{kocabasICCV2021spec} has demonstrated the ability to estimate the parameters of a perspective camera from a single image.
Adding the camera parameters to our factor graph formulation and initialising with this method may allow us to apply BodySLAM to true ``in the wild" datasets.

\clearpage

\bibliographystyle{splncs04}
\bibliography{newbib}

\clearpage
\begin{subappendices}
\renewcommand{\thesection}{\Alph{section}}%
\section{Dataset Comparison}

To fully evaluate BodySLAM, we required video sequences of human motion captured by a moving camera where camera intrinsics, ground truth camera trajectories and ground truth human body parameters were available.
As shown in Table \ref{tab:dataset_overview}, none of the existing human motion datasets include all of these required elements.
For this reason, we captured our own dataset of video sequences where the camera and human subject are both in motion, and used a motion capture system to obtain ground truth values.

\begin{table} \centering
\begin{adjustbox}{angle=-90}
{\renewcommand{\arraystretch}{1.3}%
    \begin{tabular*}{1.11\textwidth}{l @{\extracolsep{\fill}} ccccccccccc}
        & \multicolumn{11}{c}{Supplied Information} \\[2ex]
        & \rot{RGB} & \rot{stereo} & \rot{multi-view} & \rot{depth} & \rot{video} 
        & \rot{dynamic camera} & \rot{2D pose gt} & \rot{3D pose gt} 
        & \rot{not synthetic} & \rot{calibration} & \rot{\shortstack[l]{SMPL parameters\\(from MoSh)}} \\
        \hline
        \hline 
        Human3.6m \cite{h36m_pami}                    & \OK & \NO & \OK & \NO & \OK & \NO & \OK & \OK & \OK & \OK & \NO \\ \hline
        MPI-INF-3DHP \cite{mono-3dhp2017}             & \OK & \NO & \OK & \NO & \OK & \NO & \OK & \OK & \OK & \OK & \OK \\ \hline
        3DPW \cite{vonMarcard2018}                    & \OK & \NO & \NO & \NO & \OK & \OK & \OK & \OK & \OK & \OK & \OK \\ \hline
        AMASS \cite{AMASS:ICCV:2019}                  & \NO & \NO & \NO & \NO & \NO & \NO & \NO & \OK & \NO & \NO & \OK \\ \hline
        Berkeley MHAD \cite{Ofli2013BerkeleyDatabase} & \OK & \NO & \OK & \OK & \OK & \NO & \OK & \OK & \OK & \OK & \NO \\ \hline
        MPII HPE \cite{andriluka14cvpr}               & \OK & \NO & \NO & \NO & \NO & \NO & \OK & \NO & \OK & \NO & \NO \\ \hline
        Inria Stereo \cite{Ayvaci2011SparseFlow}      & \OK & \OK & \NO & \OK & \OK & \OK & \OK & \NO & \OK & \NO & \NO \\ \hline
        JTA \cite{fabbri2018learning}                 & \OK & \NO & \NO & \NO & \OK & \NO & \OK & \OK & \NO & \OK & \NO \\ \hline
        Mannequin \cite{mannequin}                    & \OK & \NO & \NO & \NO & \NO & \NO & \NO & \NO & \OK & \OK & \NO \\ \hline
        SHPED \cite{lopezquintero2015mvap}            & \OK & \OK & \NO & \NO & \OK & \OK & \OK & \NO & \OK & \NO & \NO \\ \hline
        KTP \cite{ktp:paper1,Munaro2014FastRobots}    & \OK & \NO & \NO & \OK & \OK & \OK & \NO & \NO & \OK & \OK & \NO \\ \hline
        BinoPerfCap \cite{WSVT13}                     & \OK & \OK & \NO & \NO & \OK & \OK & \NO & \NO & \OK & \OK & \NO \\ \hline
        TartanAir \cite{tartanair2020iros}            & \OK & \OK & \OK & \OK & \OK & \OK & \OK & \OK & \NO & \OK & \NO \\ \hline
        KITTI \cite{kitti:dataset}                    & \OK & \OK & \OK & \OK & \OK & \OK & \NO & \NO & \OK & \OK & \NO \\ \hline
        \hline
    \end{tabular*}}
    \end{adjustbox}
    \caption{Dataset overview}
    \label{tab:dataset_overview}
\end{table}

\pagebreak 

\section{Camera and Human Trajectory Errors}
In general, there are many possible ways of aligning the estimated camera and human centre trajectories with ground truth to compute the (average) trajectory errors, but we consider the following four as the most sensible methods:
\begin{enumerate}
    \item align the camera and human centre trajectories independently in $SE(3)$,
    \item align \emph{only} the camera trajectory in $SE(3)$ and compute the human centre trajectory based on this alignment,
    \item jointly align both human and camera trajectories in $SE(3)$, or
    \item align only the first frame of the camera poses, such that the initial condition is set to ground truth, and compute both the human and camera trajectories based off this measurement.
\end{enumerate}

In our paper, we use the first method to compute the errors, as this method allows for the fairest comparison with methods that focus only on either the camera trajectory error or the human centre trajectory error.
We use the fourth method for visualisation in the supplementary video where both trajectories must be shown simultaneously, as this method aligns the first estimates and shows the accumulating error over time.
Using other methods might confuse the viewer as the initial poses would not be aligned.

In the following sections, we provide examples of the camera and human centre trajectories estimated by BodySLAM and the baseline approach.
As discussed in the main paper, the baseline method optimises the camera trajectory via classic monocular bundle adjustment using just the camera poses and the 3D landmarks.
The human centre trajectory estimation is done using only the bundle adjusted camera poses and unary OpenPose measurements with no motion model.
Section \ref{sec:indp} shows the trajectory estimates when aligning the camera and human centre trajectories independently in $SE(3)$ (method 1), and Section \ref{sec:cam} shows the trajectory estimates when aligning only the camera trajectory in $SE(3)$ (method 2).

\pagebreak 

\subsection{Trajectories after Independent $SE(3)$ Alignment}
\label{sec:indp}
\begin{figure}[!htb]
\centering
\begin{subfigure}{.5\textwidth}
  \centering
  \includegraphics[width=.9\textwidth,trim=90 0 50 50, clip]{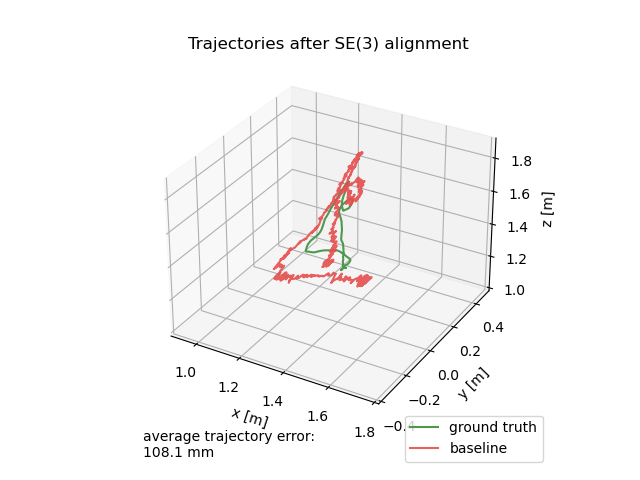}
  \caption{Baseline}
\end{subfigure}%
\begin{subfigure}{.5\textwidth}
  \centering
  \includegraphics[width=.9\textwidth,trim=90 0 50 50, clip]{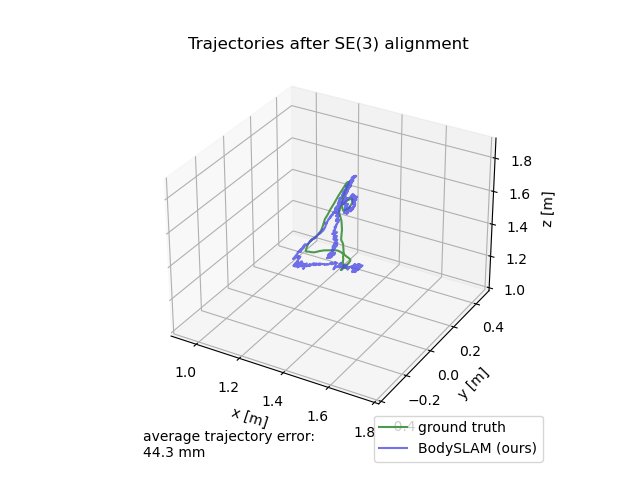}
  \caption{BodySLAM}
\end{subfigure}
\caption{Estimated camera trajectories for Sequence E2 after $SE(3)$ alignment with the ground truth camera trajectory. The baseline method optimises the camera trajectory via monocular bundle adjustment using just the camera poses and 3D landmarks. By using a human motion model to temporally constrain the optimisation problem, BodySLAM is able to accurately estimate the metric scale of the camera trajectory.}
\label{fig:ex1}
\end{figure}

\begin{figure}[!htb]
\centering
\begin{subfigure}{.5\textwidth}
  \centering
  \includegraphics[width=.9\textwidth,trim=90 0 50 50, clip]{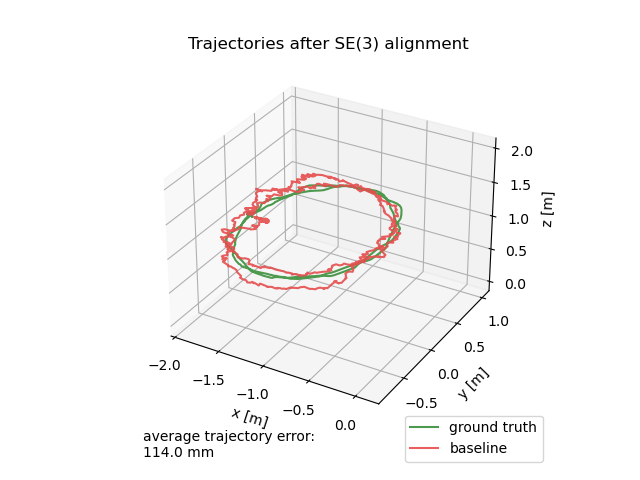}
  \caption{Baseline}
\end{subfigure}%
\begin{subfigure}{.5\textwidth}
  \centering
  \includegraphics[width=.9\textwidth,trim=90 0 50 50, clip]{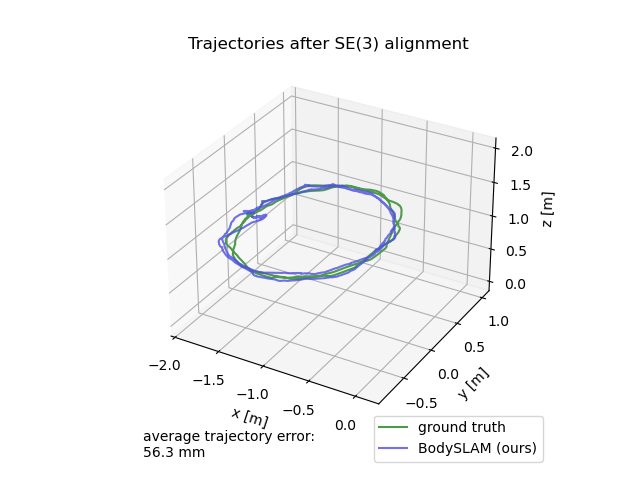}
  \caption{BodySLAM}
\end{subfigure}
\caption{Human centre trajectories for Sequence E2 after $SE(3)$ alignment with the ground truth human centre trajectory. The baseline method uses the bundle adjusted camera poses and unary OpenPose measurements to estimate human motion. In \mbox{BodySLAM}, the human centre poses and posture parameters are constrained by a motion model leading to smoother and more accurate trajectories at metric scale.}
\end{figure}
\pagebreak 

\subsection{Trajectories after $SE(3)$ Alignment of Camera Poses}
\label{sec:cam}
\begin{figure}[!htb]
\centering
\begin{subfigure}{.5\textwidth}
  \centering
  \includegraphics[width=.95\textwidth,trim=90 0 50 50, clip]{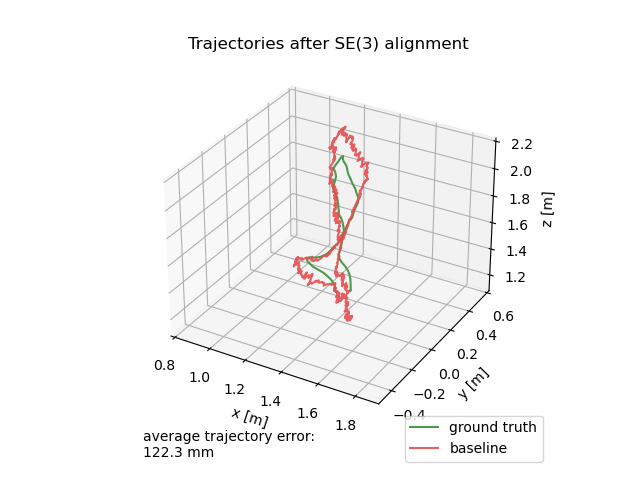}
  \caption{Baseline}
\end{subfigure}%
\begin{subfigure}{.5\textwidth}
  \centering
  \includegraphics[width=.95\textwidth,trim=90 0 50 50, clip]{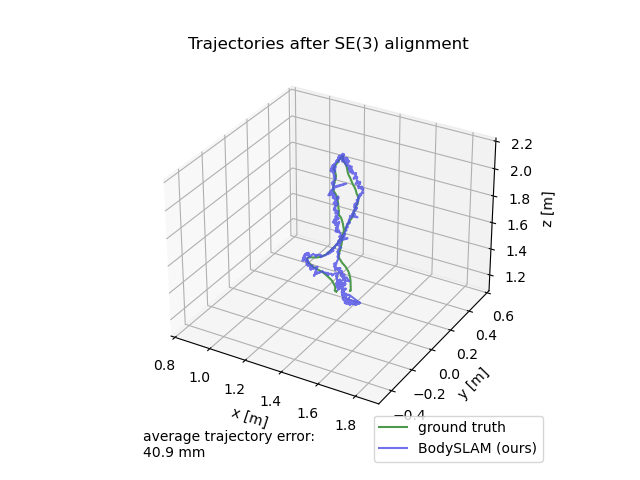}
  \caption{BodySLAM}
\end{subfigure}
\caption{Estimated camera trajectories for Sequence E4 after $SE(3)$ alignment with the ground truth camera trajectory. The baseline method optimises the camera trajectory via monocular bundle adjustment using just the camera poses and 3D landmarks. By using a human motion model to temporally constrain the optimisation problem, BodySLAM is able to accurately estimate the metric scale of the camera trajectory.}
\end{figure}

\begin{figure}[!htb]
\centering
\begin{subfigure}{.5\textwidth}
  \centering
  \includegraphics[width=.9\textwidth,trim=90 0 50 50, clip]{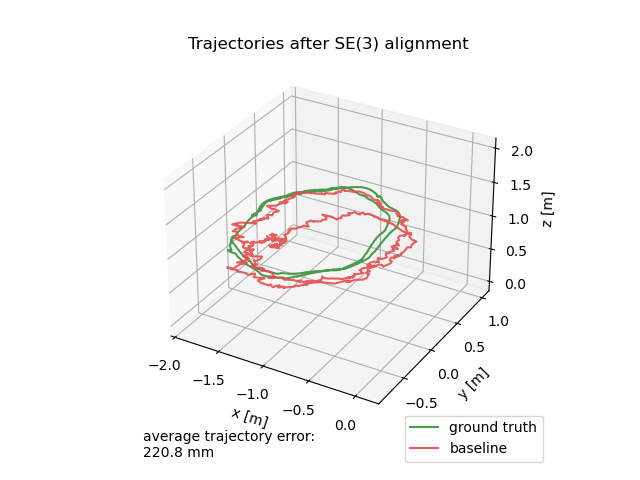}
  \caption{Baseline}
\end{subfigure}%
\begin{subfigure}{.5\textwidth}
  \centering
  \includegraphics[width=.9\textwidth,trim=90 0 50 50, clip]{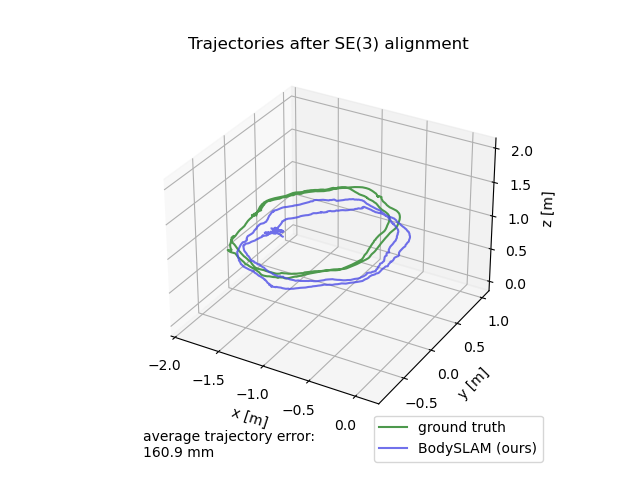}
  \caption{BodySLAM}
\end{subfigure}
\caption{Human centre trajectories for Sequence E4 after an $SE(3)$ alignment of the estimate and ground truth \textit{camera} trajectories. The baseline method uses the bundle adjusted camera poses and unary OpenPose measurements to estimate human motion. In BodySLAM, the human centre poses and posture parameters are constrained by a motion model leading to smoother and more accurate trajectories at metric scale.}
\end{figure}

\section{Joint Optimisation and Structural Landmarks}
To evaluate the influence and necessity of having a joint camera and human trajectory optimisation, we performed an ablation experiment to assess the influence of the structural landmarks to the overall performance.
In the 2-step pipeline that we used throughout our paper, where we first estimated an initial trajectory of the camera, and then added the human state to the factor graph for the joint optimisation, we introduced an additional perturbation of the camera trajectory after the first step.
This caused the average camera trajectory error to increase as seen in Table \ref{tab:struct_pert} after perturbation.
When optimising the factor graph without structural landmarks, the mean error decreased, relying only on the human keypoint measurements and the motion model error term.
However, we also show that including the structural landmarks in the second step of the optimisation, and therefore running a joint optimisation of all elements of the factor graph, improved the estimation accuracy of the camera trajectory.

We did not include the human and joint trajectory errors, since any change here in performance does only depend on the increased accuracy of the camera trajectory, rendering those results redundant.

\begin{table}[]
\centering
\caption{Optimisation of the camera trajectory with and without structural landmarks after perturbation with $\pm$ 100 mm and $\pm$ 0.01 rad. }
\begin{tabular}{l|rrrr}
\textbf{}                      & \multicolumn{4}{c}{\textbf{C-ATE {[}mm{]}}}                                                                   \\
\textbf{}                      & \multicolumn{4}{c}{SE(3)}                                                                                     \\ \hline
\multirow{2}{*}{\textbf{Seq.}} & \multirow{2}{*}{original} & \multicolumn{1}{r|}{\multirow{2}{*}{perturbed}} & optimised    & optimised        \\
                               &                           & \multicolumn{1}{r|}{}                           & no landmarks & landmarks        \\ \hline
E 2                            & 193.0                     & \multicolumn{1}{r|}{213.1}                      & 139.69       & \textbf{138.52}  \\
E 3                            & 122.3                     & \multicolumn{1}{r|}{136}                        & 78.03        & \textbf{75.56}   \\
M 4                            & 203.1                     & \multicolumn{1}{r|}{225.2}                      & 143.03       & \textbf{139.4}   \\
D 1                            & 310.4                     & \multicolumn{1}{r|}{340.7}                      & 223.08       & \textbf{219.1}   \\
D 2                            & 316.0                     & \multicolumn{1}{r|}{349.1}                      & 209.48       & \textbf{205.28}  \\ \hline
mean                           & 228.96                    & \multicolumn{1}{r|}{252.82}                     & 158.662      & \textbf{155.572}
\end{tabular}
\label{tab:struct_pert}
\end{table}

\pagebreak
\section{Scale Estimation after Perturbation}
In this ablation study, we analysed the ability of the motion model to recover the correct scale of the human and camera motion after perturbation with a scale factor.
The camera trajectory was perturbed after the initial estimation by a scale factor ranging values from $0.1$ to $5.0$.
The joint optimisation of the full factor graph was then performed, and the final optimised trajectory was aligned to ground truth in Sim(3).
The estimated scale $s$ should equal the inverse of the perturbation scale $s'$, and additionally the deviation is reported in Table \ref{tab:scale_recovery}.

\begin{table}
    \centering
    \caption{Scale estimation after perturbation. Values closer to the perturbation are better. The deviation is the estimation error relative to the perturbation. This Table contains the numerical results from Figure 4 in the paper.}
\begin{tabular}{l|rrrrrr}
\textbf{perturbation $1/s'$ {[}-{]}}      & 5.000 & 2.000 & 1.000 & 0.500 & 0.200 & 0.100 \\ \hline
\textbf{estimated scale $s$ {[}-{]}\ } & 3.673 & 2.106 & 1.051 & 0.528 & 0.224 & 0.124 \\
\textbf{deviation {[}\%{]}}        & 26.5  & 5.3   & 5.1   & 5.6   & 12.0  & 24.0 
\end{tabular}
    \label{tab:scale_recovery}
\end{table}
\end{subappendices}

\end{document}